 \documentclass[preprint, review]{elsarticle}

\usepackage{graphicx}
\usepackage{amssymb}
\usepackage{supertabular}
\usepackage{amsmath}
\usepackage{algorithmic}
\usepackage{tabularx}
\usepackage{multicol}
\usepackage{lineno}
\usepackage{booktabs}
\usepackage{lscape}
\usepackage{natbib}

\journal{Applied Soft Computing}

\begin{document}

\begin{frontmatter}

\title{An Ant Colony Optimization Algorithm for Partitioning Graphs with Supply and Demand}

\author[add1]{Raka Jovanovic}
\address[add1]{Qatar Environment and Energy Research Institute (QEERI), PO Box 5825, Doha, Qatar }

\author[add2]   {Milan Tuba}
\address[add2]	{Faculty of Computer Science, Megatrend University, Belgrade, Serbia}
\author[add3]{Stefan Vo{\ss}}
\address[add3]{Institute of Information Systems, University of Hamburg, Von-Melle-Park 5, 20146 Hamburg, Germany\\
and Escuela de Ingeniería Industrial, Pontificia Universidad Cat\'olica de Valpara\'iso, Chile}

\begin{abstract}
In this paper we focus on finding high quality solutions  for the problem of maximum partitioning of graphs with supply and demand (MPGSD). There is a growing interest for the MPGSD due to its close connection to problems appearing in the field of electrical distribution systems, especially for the optimization of self-adequacy of interconnected microgrids. We propose an ant colony optimization algorithm for the problem. With the goal of further improving the algorithm we combine it with a previously developed correction procedure.  In our computational experiments we evaluate the performance of the proposed algorithm on both trees and general graphs. The tests show that the method manages to find optimal solutions in more than 50\% of the problem instances, and has an average relative error of less than 0.5\% when compared to known optimal solutions.
\end{abstract}





\end{frontmatter}

%
\section{Introduction}

In recent years the research in the field of smart grids has had a significant increase in exploring the concept of interconnected microgrids \cite{MicroGrid}.
This approach has resulted in  novel types of typologies for electrical  grids and new aspects of such systems that should be considered.  Some of the most prominent newly emerged problems are the  maximizing of self-adequacy \cite{SelAdeq}, reliability, supply-security \cite{SupSec} and the potential for self-healing \cite{SelHea} of such systems. In many cases the underlying  optimization problems are of a very high complexity and  can not be solved to optimality in polynomial time.  Electrical   grids are  systems of gigantic size, which makes their optimization very hard from a computational point of view. Luckily, previous research has shown that for many systems it is  not necessary to use highly detailed models; often simplified graph ones can give sufficiently good approximations  to the original problem.  The family of  graph partitioning  problems  has proven to be closely related to  power supply and  delivery networks \cite{Simp1,Tree1,Simp2,Simp3,SupDem}.

In a system of interconnected microgrids each
microgrid is made as independent from the rest of the system
as possible; this results in many positive characteristics.
Some examples are the lower complexity of the entire grid and enhanced reliability of each of the microgrids due to the increased  resistance to failures in other parts of the system. The term independent is used for the case when there is a minimum of power exchange between  the connected microgrids. This property of the system is formally defined as the maximization of self-adequacy of interconnected microgrids. Recently, research has been conducted in developing algorithms for  finding approximate solutions \cite{SelAdeq} to this problem. Previous research has  also explored the  closely related problem  of efficient decomposition or islanding of large grids into islands with a balanced generation/load subject to specific constraints  \cite{sun2005simulation,li2010controlled}. Due to the large complexity and size of electrical grids, when  attempting to model and optimize some global properties, it is frequently convenient to use simplified graph models. Such models often result in different versions of graph partitioning problems suitable for specific real life applications. Some examples are  having a balanced partitioning \cite{Andreev:2004:BGP:1007912.1007931}, minimizing the number or weight of cuts \cite{Reinelt2008385,doi:10.1137/0401030}, or by limiting the number of cuts \cite{Reinelt20101}.

The focus of this paper is on the Maximal Partitioning of Graphs with Supply and Demand (MPGSD). The majority  of previous research has been dedicated to the theoretical aspects of this problem \cite{Ito2008627,Tree2,Tree1,Tree3}.  A significant part of the published work is focused on solving  this problem for specific types of graphs like trees \cite{Tree2,Tree1,Tree3} and series-parallel graphs   \cite{Ito2008627}. A method for finding solutions with a guarantee of a $2k$-approximation for general graphs has been presented in \cite{SupDem}. Different variations of the original problem have been developed, like a parametric version \cite{Parametric} and one with  additional capacity constraints \cite{Tree4}.

In this paper we present an ant colony optimization (ACO) \cite{dorigo2005ant} approach for finding high quality approximate solutions to the MPGSD. ACO has previously been successfully applied to problems of   multiway \cite{ilc2011distributed} and balanced \cite{Balanced} graph partitioning. The same method has also proven to be suitable for the   closely related problems of graph cutting \cite{Cutting} and covering \cite{Jovanovic20115360,Raka2} and partitioning of meshes \cite{Mesh}. The proposed ACO adaptation for our problem of interest is based on the greedy algorithm presented in \cite{BasicAlgorithm,MultiHeuristicGSD}. The ACO algorithm is further  improved by combining it with our previously developed correction procedure \cite{MultiHeuristicGSD}. In our tests on general graphs and trees, we show that the newly developed method frequently manages to find optimal  solutions and has a small average error when compared to known optimal solutions.
	
The paper is organized as follows. In the second section we give the definition of the MPGSD. Then we provide a short outline of a greedy algorithm which is used as a basis for the proposed method. In the third  section we present details of our ACO algorithm. In the subsequent sections we discuss results of our computational experiments and provide some conclusions.

\section{Maximal Partitioning  of  Graphs With Supply/Demand}


The MPGSD is defined for an undirected graph  $G=(V,E)$  with a set of nodes $V$ and a set of edges $E$. The set of nodes $V$ is split into two disjunct subsets $V_s$  and $V_d$. Each node $u\in V_s$ will be called a supply vertex and will have a corresponding positive integer value $sup(u)$. Elements of the second subset $v \in V_d$ will be called demand vertices and  will have a corresponding positive integer value $sup(v)$.   The goal is to find a partitioning $\Pi= \{S_1, S_2,.., S_n\}$ of the graph $G$ that satisfies the following constraints.  All the subgraphs in $\Pi$ must be connected subgraphs  containing  only  a single distinct supply node. As a result we have $|V_s| = n$.  Each of the $S_i$  must have a supply greater or equal  to its total demand. Each demand vertex can be an element of only one subgraph, or in other words it can only receive 'power' from one supply vertex through the edges of $G$.

With the intention of having a  simpler notation, we will use strictly  negative values for demand nodes and positive values for supply nodes (note that this is slightly different from the definition of \cite{Ito2008627}).
The goal is to maximize the fulfillment of demands, or more precisely to maximize the following sum.
\begin{equation}
D(\Pi) = - \sum_{S \in \Pi}\sum_{v \in S\cap V_d}sup(v)
\end{equation}
while the following constraints are satisfied for all $S_i \in \Pi$
\begin{eqnarray}
\label{constarint}
\sum_{v \in S_i}sup(v) \geq 0 \\
\label{constarint2}
S_i \cap S_j = \emptyset  \,\,\,\,\, , \,\, i\neq j\\
S_i \,\,\,\, is \,\,\,\,connected
\end{eqnarray}

\begin{figure}[tcb]
\centering
\includegraphics[width=0.85\textwidth]{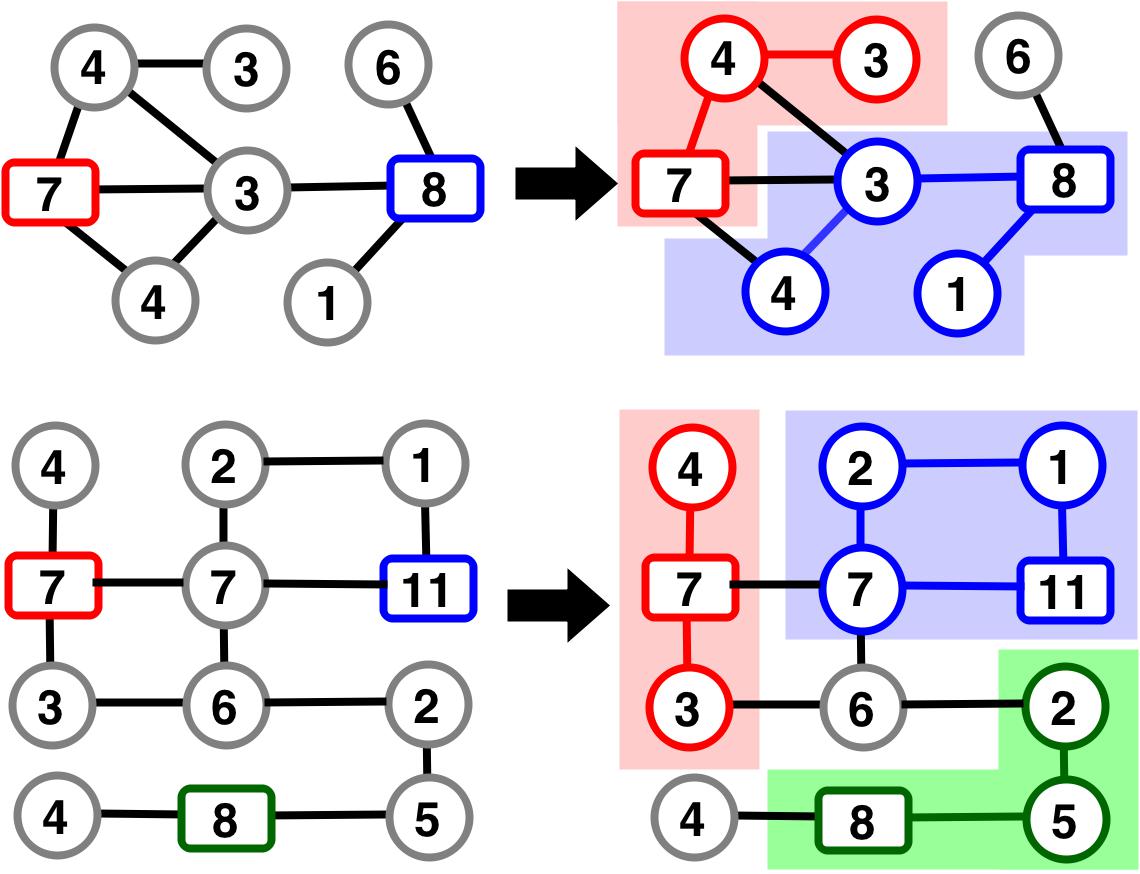}
\caption{Examples of problem instances for the MPGSD. On the left the  square nodes represent supply nodes and circles demand nodes. Numbers within the nodes correspond to supply and demand values, respectively. The right side shows the solutions, where the same color (or connected shaded set) of nodes indicates they are a part of the same partitioning.}
\label{fig:ProblemColor}
\end{figure}

An illustration of the MPGSD is given in Figure \ref{fig:ProblemColor}.
 It has been shown that the MPGSD is NP-hard even in the case of a graph containing  only one supply node and having a star structure \cite{Ito2008627}.

\section{Outline of Greedy Algorithm}

In this section we give a short overview of the greedy algorithm, for which details  can be found in \cite{BasicAlgorithm}, that is used as a base for the ACO method for the MPGSD.  As previously stated, the solution  of the problem of interest is a set of $|\Pi|=n$ subgraphs where $n=|V_s|$ is the number of supply nodes.  In the initial step of the algorithm we start with $n$ disjunct subgraphs $S_i$, that  only contain one supply node $S = \{s_i \}$. At each of the following steps (iterations) of the algorithm  one vertex $v \in V_d$ is selected and used to expand a selected subgraph  $S_i$. The selection of both $v$ and $S_i$ is performed in a way that the newly generated subgraph satisfies the constraints of being connected, disjunct and fulfills Eq. \ref{constarint}.

 Let us define $NV$  as the set of adjacent vertices to $v$ in $G$ using the following equation
\begin{equation}
	NV(v) = \{u| u \in V \wedge (u,v) \in E\}
\end{equation}

The idea is to gradually expand each of the subgraphs $S_i$ by adding new vertices $v$ to them. The set of potential candidates for expansion of subgraph  $S_i$, or in other words vertices that are adjacent to the $S_i$, can be defined using the extension of $NV$ to subgraphs. It is important to note that as the subgraph $S_i$ will be changed in subsequent iteration, the notation $S_i^k$ will be used to specify the state of subgraph $S_i$ at iteration $k$. Now we can extend the definition of $NV$ with the following equation.
\begin{equation}
    \label{NV_SG}
	\hat{N_i^k} = NV(S_i^k) = \{u| u \in V \wedge \exists (v \in S_i^k) (u,v) \in E\}
\end{equation}

It is evident that if the expansion of subgraph $S_i^k$ is done by $v \in \hat{N_i^k}$  the newly created subgraph $S_i^{k+1}$ will be connected, but it is not necessary that the other constraints will be satisfied. More precisely,  the new $S_i^{k+1}$ need not satisfy  Eq. \ref{constarint},  or there may exist such an $S_j^{k+1}$ that   $S_j^{k+1} \cap S_i^{k+1} = v$ for the new subgraph.  This can be avoided if instead of using set $\hat{N_i^{k}}$, we use a restricted set of vertices $N_i^k$ that guarantees that the constraints will be satisfied when $S_i^{k}$ is expanded using $v \in N_i^k$. We shall first define  $sup_i^k$ as the available supply for subgraph $S_i$ at iteration $k$ in the following equation.
\begin{equation}
    \label{Available}
	 sup_i^k = \sum_{v \in S_i^k}sup(v)
\end{equation}
Using $sup_i^k$ given in Eq. \ref{Available}, $N_i^k$ is defined in the following way.
\begin{equation}
    \label{NV_SG}
	N_i^k = \{u| u \in \hat{N_i^k} \wedge sup(u) \leq sup_i^k\} \setminus \bigcup^n_{j=1}S_j^k
\end{equation}

The sets $N_i^k$ are used to  specify the  greedy algorithm for the MPGSD in combination with two heuristic functions. More precisely, at each iteration one    heuristic $hs$ is used to select the subgraph $S_i^k$ most suitable for expansion, and the second heuristic $hv$ will be used to select the best $v \in N_i^k$ to be added to $S_i^k$. An extensive analysis of potential heuristics is given in our previous work given in articles\cite{BasicAlgorithm,MultiHeuristicGSD}. This procedure will be repeated until it is not possible to expand any of the subgraphs.

\section{Application of Ant Colony Optimization}

In this section we present an ACO approach  for solving the MPGSD, based on the greedy algorithm from \cite{BasicAlgorithm}, outlined in the previous section. The general idea of ACO algorithms is to perform an "intelligent" randomization of an appropriate greedy algorithm for the problem of interest. There are several variations of ACO, out of which the  Ant Colony System \cite{dorigo1997ant} is most commonly used.   In it the "intelligence" comes from experience gained  by previously generated solutions, which is stored in a pheromone matrix.  In practice, a colony of $n$ artificial ants generates  solutions using a probabilistic algorithm based  on a heuristic function and the pheromone matrix.  As in the case of the greedy algorithm, an ant generates a solution by expanding a partial one through steps. The difference is that instead of using a heuristic function it uses a probabilistic transition rule to decide what is to be added to the partial solution. The pheromone matrix stores the experience gathered by the artificial ants. This is done by applying a global and local update rule to the pheromone matrix.  The global update rule is used after all $n$ ants in the colony have generated a solution and it reinforces the selection of elements inside of the best found solution or in some variations of good solutions.  The local update is performed after an ant has applied the transition rule, and its purpose is to diversify the search of the solution space by avoiding the selection of the same elements of the solution by all of the ants.

Before defining the ACO algorithm for the MPGSD, we will first state some observations regarding the greedy algorithm and the form of the solution of the problem. A solution $\Pi$ of the MPGSD can also be observed as a set of pairs $(s, v)$, which states that node $v$ is inside subgraph $S_s$. In this notation we will include $(-1,v)$ for the case where $v$ is not a member of any subgraph $S_s$. From this type of notation we realize that in the algorithm given in the previous section only the second stage, the selection of node $v$, directly specifies the elements of the solution. The purpose of the heuristic in the first stage is to make it possible to perform a good expansion of the partial solution, which is of significant importance when only one solution is generated using a deterministic algorithm. In case of an ACO algorithm this becomes less important since many solutions are generated and the "steering" in the direction of good solutions is, to a large extent, done by the pheromone matrix.

Because of this, in the proposed ACO algorithm the heuristic function at this stage will be substituted with a random selection from the set of subgraphs that can be expanded. In this way the ACO mechanism will only be dedicated to the selection of expansion nodes.

\subsection{Algorithm Specification}

To specify the ACO for the MPGSD  we need to define the transition rule, global  and local update rules. In all of the following equations we will assume that we have a randomly selected subgraph $S_s$ with index $s$.  We will first define the transition rule, based on the same heuristic function as in \cite{BasicAlgorithm}, defined in the following equation.
\begin{equation}
    \label{HV}
	\eta_v = hn_1(v) = |sup(v)|
\end{equation}

The heuristic function $\eta_v$ given in Eq. \ref{HV} states that vertices with high demand are considered more desirable. The logic behind this is that it gets harder to satisfy high demands as the algorithm progresses since the available supply decreases as new vertices are added to the subgraphs. Because of this it is better to resolve high demands early.

Using $\eta_v$ we can define the transition rule for individual ants. This selection is done using a combination of deterministic and probabilistic steps. First we need to include the constraint  that  only  vertices from the set $N_i^k$ can be selected. We specify this constraint using the following equation.
\begin{equation}
\label{DefinicijaRaspodele}
   p_v^k  = \left\{ \begin{array}{l}
            0 \,\,\,\,\,\,\,\,\,\,\,\,\,\,\,\,\,\,\,\,\,\,\, ,v \not \in N_s^k\\
            prob_i^k  \,\,\,\,\,\,\,\,\,\,\, ,v \in N_s^k\\
 \end{array} \right.
\end{equation}

In Eq. \ref{DefinicijaRaspodele}, $p_v^k$ gives us the probability of selecting node  $v$  at step $k$. As previously stated we only consider $v \in  N_s^k$ where $s$ is the selected subgraph, as a consequence the probability of selecting $v \notin N_s^k$ is $0$. For the nodes that are elements of $N_s^k$ their selection is done using the following formula.
\begin{equation}
\label{DefinicijaProbability}
  prob_v^k  = \left\{ \begin{array}{l}
 1\,\,\,\,\,\,\,\,\,\,\,\,\,\,\,\,\,\,\,\,\,\,\,\,\,\,\,\,\,\,\,\,\,\,\,\,\,\,,q > q_0 \,\,\, \&  \,\,\, v = \arg \mathop {\max }\limits_{i \in N_s^k } \tau _{is} \eta _i   \\
 0\,\,\,\,\,\,\,\,\,\,\,\,\,\,\,\,\,\,\,\,\,\,\,\,\,\,\,\,\,\,\,\,\,\,\,\,\,\,,q > q_0 \,\,\, \&  \,\,\, v \ne \arg \mathop {\max }\limits_{i \in N_s^k } \tau _{is} \eta _i   \\
  \frac{{\tau _{vs} \eta _v  }}{{\sum\nolimits_{i \in N_s^k} {\tau _{is} \eta_i } }}\,\,\,\,\,\,\,\,\,\,\,\,\,\,\,\,\,\,\,\,\,\,,q \le q_0  \\
 \end{array} \right.
\end{equation}

In Eq. \ref{DefinicijaProbability} $prob^k_v$ gives us the probability of selecting node $v$ at step $k$. The values of the pheromone matrix $\tau _{is}$ correspond to elements of the solution in the form of  a vertex-subgraph pair $(i,s)$.  In the same equation parameter $q_0$ is used to define the exploitation/ exploration rate. Connected to it, $q \in (0,1)$ is a  random variable which specifies whether the next selected node will be deterministic or non-deterministic.  In the case of the former $q>q_0$, we simply select the node $v$ with the maximal value of
$\tau _{is} \eta _{i}$, which results in a probability $1$. If the selection is  non-deterministic ($q<q_0$), the probability distribution for node selection is given in the last row of Eq. \ref{DefinicijaProbability}.

The next component of the ACO method  that needs to be specified is the global update rule. The proposed ACO algorithm is based on the ant colony system, in which only the best found solution deposits pheromone after each iteration of the colony. This update is formally defined using the following equations
\begin{equation}
\label{DefinicijaDeltaTau} \Delta\tau  =  Val(\Pi')
 \end{equation}

 \begin{equation}
\label{DefiniciajTau} \tau _i  = (1 - p)\tau _{vs}  + p \Delta\tau  \,\,\,\,\,\,\,\,\,\,\,\,\, ,  \forall(v,s)           \in \Pi'
 \end{equation}

In Eq. \ref{DefinicijaDeltaTau} $\Pi'$ is used to note the currently  best found solution. $\Delta\tau$ is used to specify the quality of the solution $\Pi'$ using function $Val$ for which we will give details in the implementation subsection. In Eq. \ref{DefiniciajTau}, the parameter $p \in (0,1)$   is used to specify the influence of the global update rule. It is important to point out that Eq. \ref{DefiniciajTau} only effects the  values of pheromone $\tau _{vs}$ for  $(v,s) \in \Pi'$.

As previously mentioned the local update rule is applied after individual ants perform the transition rule. In our implementation the local update rule is applied after an ant $i$ has generated a solution $\Pi_i$ using the following formula
 \begin{equation}
 \label{DefinicijaLocal}
 \tau_{vs}  = \varphi \tau_{vs} \,\,\,\,\,\,\,\,\,\,\,\,\, ,\forall(v,s)           \in \Pi_i
  \end{equation}

In Eq. \ref{DefinicijaLocal} $\varphi \in (0,1)$ is used to specify the influence of the local update rule.

\subsection{Implementation}

In this section we give details of the implementation of the proposed ACO algorithm. The first necessary step is to define a suitable quality function $Val$ for the generated solutions. This is done by using the following equations.
 \begin{eqnarray}
  T =  \sum_{v \in V_s} sup(v)\\
 \label{Quality}
  Val(\Pi) = \frac{1}{T - D(\Pi)+1}
  \end{eqnarray}

 Eq.~\ref{Quality} states that the quality of the solution will be inversely proportional to the difference of $T$, the total initially available supply in $G$ and the satisfied demand $D(\Pi)$ of partitioning $\Pi$. To avoid division by zero one is added to this value. Using this measure, the initial value of all the pheromone matrix elements $t_{vs}$ is set to the value $Val(\Pi_g)$, where $\Pi_g$ is the solution acquired using the previously  outlined greedy algorithm. More precisely, it corresponds to the method presented in \cite{BasicAlgorithm}, where the node selection heuristic is the maximal demand and the subgraph selection heuristic is the maximal available supply.

With the goal of having a  better presentation  of the proposed method, it is presented in the form of the following pseudo-code

\begin{algorithmic}
\STATE Generate solution $\Pi_g$ using the greedy algorithm
\STATE Initialize the pheromone matrix $\tau$ with $Val(\Pi_g)$
 \WHILE{(Maximal number of iterations not reached)}
  \FORALL{$n$ ants}
  	     \STATE{ Initialize $\Pi = \{S_1,..,S_n \}$ , $S_i = \{s_i\}$}
   \WHILE{$\Pi$ can be expanded}
	     \STATE{Randomly select $S$, where $|NV(S)|>0$}
	     \STATE{Select $v$ for $S$ using transition rule}
	     \STATE $\Pi = \Pi \cup (v,S)$
		 \STATE{Update auxiliary structures}
	 \ENDWHILE
		 \STATE{Apply correction procedure to $\Pi$}
	     \STATE{Apply local update rule for $\Pi$}

  \ENDFOR
 	     \STATE{Apply global update rule for $\Pi_{best}$}
 \ENDWHILE
\end{algorithmic}

As illustrated in the pseudo-code, the first step is generating a solution using a greedy algorithm and   initializing the pheromone matrix $\tau$. The main loop performs one iteration for the colony of ants by generating a solution for each of the $n$ artificial ants. For each of the ants we start with the initial partitioning $\Pi$. At each iteration of the following loop, we randomly select a subgraph $S$, and using the transition rule a node $v$ is selected for expansion. After each such step it is necessary to update the auxiliary structures, presented in \cite{BasicAlgorithm}, that are used to make the proposed algorithm  computation efficient.

After an ant has generated a solution $\Pi$ we apply the correction procedure, presented in \cite{MultiHeuristicGSD}, which corresponds to a local search to improve its quality. This is done due to the fact that, in general, ACO algorithms  have a problem with narrowing on  local minima.  It has been shown that the performance of such methods can be significantly improved if they are combined with a local search. For the newly acquired solution we apply the local update rule given in Eq.~\ref{DefinicijaLocal}. After all of the ants in the colony have generated their solutions we apply the global update rule  given in Eq.~\ref{DefiniciajTau} for the best solution $\Pi_{best}$ found by the algorithm for all the previous iterations.

\section{Results}

In this section we present the results of our computational experiments used to evaluate the  performance of the proposed ACO methods. We give a comparison of  the proposed ACO algorithm, with (ACO-C) and without (ACO) the use of a correction procedure, and the basic greedy algorithm (Gr). All the algorithms  have been implemented in C\# using Microsoft Visual Studio 2012. The source code and the executive files have been made available at \cite{Data}. The calculations have been done on a machine with Intel(R) Core(TM) i7-2630 QM CPU \@ 2.00 Ghz, 4GB of DDR3-1333 RAM, running on Microsoft Windows 7 Home Premium 64-bit.

\begin{table*}[htb]
\footnotesize
\center
\caption{\label{table:GenCor}Comparison of the proposed algorithms for general graphs  when 1500 solutions have been generated for ACO and ACO-C.}
\begin{tabularx}{410pt}{X*{13}{c}}

\toprule
Sup X Dem&  \multicolumn{3}{c}{$Avg(Stdev)$}  & \multicolumn{3}{c}{$Max$} & \multicolumn{3}{c}{$Hits$}\\

         & Gr &ACO & ACO-C&  Gr &ACO & ACO-C & Gr &ACO & ACO-C \\
\midrule
2 X 6 & 7.45(8.71) & 0.28(1.77) & 0.00(0.00) & 46.10 & 11.36 & 0.00 & 17 & 39 & 40 \\
2 X 10 & 5.62(4.39) & 0.24(0.60) & 0.00(0.00) & 23.08 & 3.04 & 0.00 & 4 & 32 & 40 \\
2 X 20 & 1.85(1.11) & 0.09(0.13) & 0.00(0.00) & 4.21 & 0.46 & 0.00 & 1 & 26 & 40 \\
2 X 40 & 0.77(0.50) & 0.00(0.00) & 0.00(0.00) & 1.74 & 0.00 & 0.00 & 3 & 40 & 40 \\
\midrule
5 X 15 & 10.88(7.77) & 0.59(1.31) & 0.13(0.44) & 38.78 & 7.14 & 2.22 & 0 & 28 & 36 \\
5 X 25 & 7.89(5.99) & 0.78(0.77) & 0.22(0.29) & 34.62 & 3.84 & 1.07 & 0 & 7 & 21 \\
5 X 50 & 3.89(2.62) & 0.15(0.13) & 0.01(0.03) & 10.27 & 0.65 & 0.10 & 0 & 8 & 35 \\
5 X 100 & 2.01(2.54) & 0.02(0.03) & 0.00(0.00) & 13.63 & 0.13 & 0.00 & 0 & 26 & 40 \\
\midrule
10 X 30 & 11.53(4.44) & 0.51(0.82) & 0.16(0.40) & 23.88 & 4.29 & 1.60 & 0 & 19 & 32 \\
10 X 50 & 7.36(2.79) & 1.08(0.45) & 0.26(0.26) & 14.19 & 2.20 & 0.90 & 0 & 0 & 13 \\
10 X 100 & 3.92(2.44) & 0.28(0.14) & 0.05(0.05) & 13.14 & 0.69 & 0.18 & 0 & 0 & 18 \\
10 X 200 & 2.52(2.81) & 0.10(0.05) & 0.00(0.00) & 12.98 & 0.22 & 0.00 & 0 & 1 & 40 \\
\midrule
25 X 75 & 12.14(3.16) & 1.63(1.08) & 0.28(0.29) & 19.23 & 5.38 & 1.14 & 0 & 1 & 12 \\
25 X 125 & 8.64(2.07) & 1.76(0.58) & 0.51(0.31) & 13.64 & 3.16 & 1.49 & 0 & 0 & 0 \\
25 X 250 & 4.60(1.49) & 0.83(0.19) & 0.13(0.06) & 8.68 & 1.22 & 0.23 & 0 & 0 & 0 \\
25 X 500 & 2.81(1.37) & 0.44(0.07) & 0.01(0.02) & 6.10 & 0.56 & 0.06 & 0 & 0 & 11 \\
\midrule
50 X 150 & 12.04(1.86) & 2.20(0.78) & 0.46(0.40) & 15.63 & 3.91 & 1.78 & 0 & 0 & 3 \\
50 X 250 & 8.76(1.34) & 2.67(0.47) & 0.84(0.22) & 10.80 & 3.59 & 1.42 & 0 & 0 & 0 \\
50 X 500 & 4.65(1.28) & 1.56(0.23) & 0.31(0.07) & 7.39 & 2.27 & 0.50 & 0 & 0 & 0 \\
50 X 1000 & 3.07(0.99) & 0.73(0.11) & 0.06(0.02) & 5.97 & 0.99 & 0.13 & 0 & 0 & 0 \\
\midrule
100 X 300 & 11.75(1.45) & 3.69(0.69) & 0.90(0.42) & 14.61 & 6.05 & 2.02 & 0 & 0 & 0 \\
100 X 500 & 8.77(1.07) & 3.93(0.60) & 1.42(0.28) & 11.65 & 6.21 & 2.13 & 0 & 0 & 0 \\
100 X 1000 & 4.67(0.89) & 2.29(0.22) & 0.60(0.07) & 7.04 & 2.71 & 0.74 & 0 & 0 & 0 \\
100 X 2000 & 3.04(0.73) & 1.11(0.22) & 0.14(0.04) & 4.58 & 1.82 & 0.27 & 0 & 0 & 0 \\
\bottomrule

\end{tabularx}
\end{table*}

\begin{table*}[htb]
\footnotesize
\center
\caption{\label{table:TreeCor}Comparison of the proposed algorithms for trees when 1500 solutions have been generated for ACO and ACO-C. }
\begin{tabularx}{410pt}{X*{13}{c}}

\toprule
Sup X Dem&  \multicolumn{3}{c}{$Avg(Stdev)$}  & \multicolumn{3}{c}{$Max$} & \multicolumn{3}{c}{$Hits$}\\

         & Gr &ACO & ACO-C&  Gr &ACO & ACO-C & Gr &ACO & ACO-C \\
\midrule
2 X 6 & 1.67(5.92) & 0.00(0.00) & 0.00(0.00) & 26.37 & 0.00 & 0.00 & 37 & 40 & 40 \\
2 X 10 & 5.46(8.02) & 0.11(0.50) & 0.02(0.13) & 35.14 & 3.08 & 0.85 & 16 & 37 & 39 \\
2 X 20 & 8.71(9.11) & 0.09(0.34) & 0.01(0.07) & 28.94 & 2.13 & 0.43 & 3 & 35 & 39 \\
2 X 40 & 6.09(7.65) & 0.05(0.14) & 0.00(0.00) & 30.58 & 0.57 & 0.00 & 3 & 34 & 40 \\
\midrule
5 X 15 & 8.47(8.71) & 0.01(0.04) & 0.00(0.00) & 27.19 & 0.27 & 0.00 & 13 & 39 & 40 \\
5 X 25 & 7.87(6.04) & 0.10(0.25) & 0.07(0.29) & 21.80 & 1.11 & 1.49 & 1 & 33 & 37 \\
5 X 50 & 10.63(6.99) & 0.07(0.16) & 0.04(0.15) & 29.60 & 0.89 & 0.89 & 0 & 28 & 35 \\
5 X 100 & 16.43(11.22) & 0.12(0.64) & 0.00(0.00) & 50.93 & 4.09 & 0.00 & 0 & 30 & 40 \\
\midrule
10 X 30 & 8.66(6.44) & 0.09(0.25) & 0.01(0.06) & 27.17 & 1.13 & 0.37 & 2 & 34 & 39 \\
10 X 50 & 9.67(5.59) & 0.07(0.17) & 0.07(0.21) & 29.53 & 0.70 & 1.08 & 0 & 31 & 34 \\
10 X 100 & 11.40(6.33) & 0.09(0.13) & 0.03(0.09) & 26.73 & 0.53 & 0.48 & 0 & 19 & 33 \\
10 X 200 & 13.92(6.58) & 0.27(1.15) & 0.25(1.15) & 26.02 & 6.71 & 6.71 & 0 & 23 & 37 \\
\midrule
25 X 75 & 9.52(4.87) & 0.18(0.35) & 0.03(0.12) & 22.49 & 1.44 & 0.73 & 0 & 26 & 36 \\
25 X 125 & 10.79(3.83) & 0.15(0.16) & 0.06(0.13) & 17.29 & 0.63 & 0.47 & 0 & 12 & 27 \\
25 X 250 & 10.68(3.22) & 0.29(0.60) & 0.06(0.23) & 20.23 & 2.73 & 1.31 & 0 & 9 & 30 \\
25 X 500 & 11.64(3.93) & 0.48(0.69) & 0.14(0.34) & 19.37 & 2.72 & 1.27 & 0 & 2 & 30 \\
\midrule
50 X 150 & 8.66(2.93) & 0.15(0.18) & 0.04(0.09) & 17.04 & 0.76 & 0.46 & 0 & 13 & 30 \\
50 X 250 & 10.20(3.07) & 0.31(0.29) & 0.07(0.09) & 18.72 & 1.23 & 0.39 & 0 & 2 & 17 \\
50 X 500 & 11.92(3.03) & 0.44(0.50) & 0.05(0.13) & 18.84 & 2.21 & 0.79 & 0 & 0 & 11 \\
50 X 1000 & 12.75(2.41) & 1.09(0.85) & 0.51(0.60) & 18.30 & 3.61 & 1.92 & 0 & 0 & 10 \\
\midrule
100 X 300 & 9.83(1.99) & 0.27(0.18) & 0.09(0.15) & 14.11 & 0.81 & 0.64 & 0 & 2 & 17 \\
100 X 500 & 10.26(1.79) & 0.56(0.35) & 0.08(0.06) & 14.43 & 1.62 & 0.21 & 0 & 0 & 3 \\
100 X 1000 & 11.18(1.82) & 1.05(0.51) & 0.18(0.29) & 14.55 & 2.25 & 1.55 & 0 & 0 & 3 \\
100 X 2000 & 12.07(1.86) & 2.03(0.75) & 0.97(0.75) & 17.49 & 3.69 & 3.99 & 0 & 0 & 0 \\

\bottomrule

\end{tabularx}
\end{table*}

To have an extensive  evaluation of the proposed algorithm tests have been conducted on a wide range of graphs. We have used 24 different graph sizes having  2-100 supply nodes and 6-2000 demand nodes.  For each of the test sizes 40 different problem instances have been generated. We have compared the average solution and the number of found optimal solutions for each size. With the goal of observing the potential dependence between the method performance and graph structure, we have performed tests on trees and general graphs. The same data sets have been used in the article \cite{MultiHeuristicGSD}, which can be downloaded from \cite{Data}, where specifics of the method for their generation are presented. It is important to note that the optimal solutions are known for each of the test instances due to the algorithm used for their generation.

For each of the 40 problem instances, inside of one graph size, a single run of the ACO algorithm has been performed for both versions of the method.  In each of the runs the colony had $10$ ants and 150 iterations have been performed. In practice this means that 1500 solutions have been generated for each test instance. The  parameters for specifying  the influence of the global and the local update rules had the following values  $p=0.1$ and $\varphi=0.9$. We have used the value  $q_0=0.1$ to define the exploitation/exploration rate. The chosen parameters correspond to the commonly used values for the ACO algorithm, for which our initial tests have shown that they give the best performance.

The results of the conducted computational experiments are presented in Tables \ref{table:GenCor}, \ref{table:TreeCor} for general graphs and trees, respectively.  The values in these tables represent the average normalized error of the found solutions compared to the known optimal one, for each of the used methods. More precisely, for each of the 40 test instances, for each graph size,  the normalized error is calculated by $(Optimal - found)/Optimal *100$, and we show the average value in  Tables \ref{table:GenCor}, \ref{table:TreeCor}. To have a better comprehension of the performance we have also included the standard deviation and maximal errors. The last value included in these tables is the number of found optimal solutions (hits) for each graph size out of the 40 test instances.

For both types of graphs, there is a very significant improvement  of the ACO algorithms when compared to the basic greedy algorithm. In case of the greedy algorithm the average error varies from less than 1\% to even 16\%, while for the ACO algorithm the range is between 0-4\%, and in case of ACO-C  it is within 0-1.5\%.  The ACO-C method proves to be very robust in the sense that  maximal error has never  exceeded 7\%, and has been greater than 1\% in only 17 out of the 48 graphs sizes.  The basic ACO  significantly improves the maximal error when compared with the greedy algorithm but lacks behind ACO-C with the maximal error never exceeding 12\% and being less than 1\% in 50\% of test sizes.

It is important to note that there is a difference in performance of the methods for general graphs and      trees. While the greedy algorithm has a slightly better performance in case of general graphs, in case of the proposed ACO algorithms we have an opposite situation. In case of trees, the ACO  had only twice an average error higher than 1\% and never higher than 2.03\%. ACO-C produces even better results with never having an error greater than 1\%, and having an error of less than 0.1\% in 19 out of 24 graph sizes.

The results in Tables \ref{table:GenCor} and \ref{table:TreeCor} show that the greedy algorithm only manages to find optimal solutions in the case of the smallest graphs. On the other hand the  ACO-C manages to find the optimal solution for about 50\% of the test instances, while ACO is close to 30\%. As in the case of average errors both ACO methods have a significantly better performance for trees than for general graphs. In case of trees ACO-C has found the optimal solution for 65\% of the test instances, but it would rarely find ones for the largest graphs.

\begin{figure}[tcb]
\centering
\includegraphics[width=0.85\textwidth]{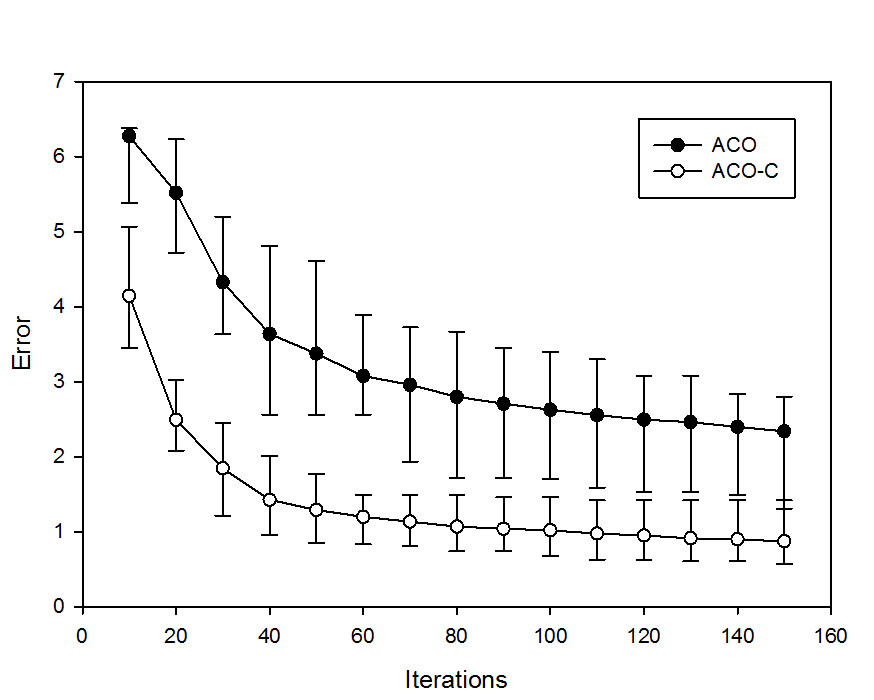}
\caption{Illustration of performance of ACO and ACO-C for 20 runs for a general graph with 50 supply nodes and 250 demand nodes. The graph shows the average error for the 20 runs at each iteration. The range represents the maximal and minimal error at each step.}
\label{fig:General}
\end{figure}

\begin{figure}[tcb]
\centering
\includegraphics[width=0.85\textwidth]{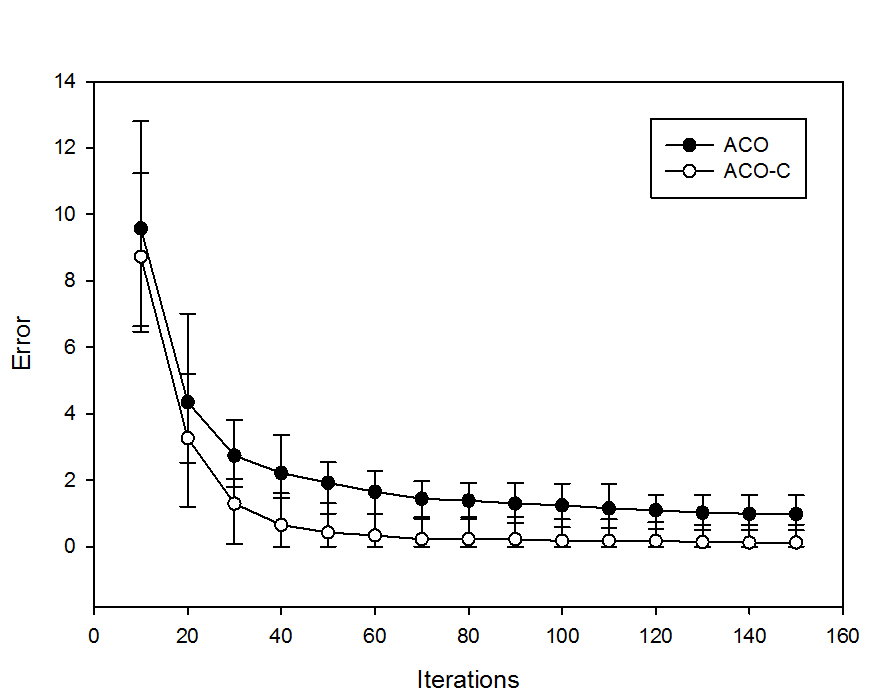}
\caption{Illustration of performance of ACO and ACO-C for 20 runs for a tree with 50 supply nodes and 1000 demand nodes. The graph shows the average error for the 20 runs at each iteration. The range represents the maximal and minimal error at each step.}
\label{fig:Tree}
\end{figure}

Due to the stochastic nature of the ACO algorithm, we have also performed multiple runs of ACO and ACO-C, with different seeds for the random number generator, on a single problem instance. In case of this type of analysis, for small problem instances both methods have a very good performance and manage to find the best solution for the vast majority of runs. The interesting cases are connected to the -- harder to solve -- large graph sizes. The behavior of both algorithms is similar for all tested instances in one problem size, because of which we believe it is sufficient to give a graphic illustration only for a single test instance for general graphs and trees in Figures \ref{fig:General}, \ref{fig:Tree}. In case of general graphs, in Figure \ref{fig:General}, we can see that ACO-C has a significantly higher speed of convergence. ACO-C also proves to be significantly more reliable, since its maximal error is lower than the minimal error for ACO. We can also see that both methods have a notable dependence on the selected seed of the random number generator since the difference between minimal and maximal error is 2\% and 1\% for ACO and ACO-C, respectively. The results in Figure \ref{fig:Tree}, for trees, show that the advantage of ACO-C is much lower when compared to ACO. The range of error for both methods is much lower in case of trees.

From  the performed tests we can conclude that the proposed ACO algorithm is a very effective method for solving the MPGSD. The tests have also shown that, as for many others, in the case of the problem of interest the performance of the ACO algorithm can be significantly improved by adding a local search method. Finally, the quality of the solutions acquired by the ACO and ACO-C is to a certain extent dependent on the seed of the random number generator. Because of this fact, when applying the proposed method it is advisable to perform multiple runs to get the highest quality of found solutions.

\section{Conclusion}

In this paper we have presented an ant colony optimization algorithm for solving  the problem of the  maximum partitioning of graphs with supply and demand. To the best of our knowledge,  this is the first time that the ACO metaheuristic  has been applied to this type of problem.  The basic ACO algorithm has been combined with a local search to enhance the performance of the method.  Our computational experiments have shown that the proposed approach managed to find the optimal solutions in more than 50\% of the test problem instances, and had an average relative error of less then 0.5\%. The tests have been performed on trees and general graphs and shown that the method is more suitable for trees.

In the future we plan to adapt the method to a less constrained and a stochastic version of the problem. This type of research can prove to be very beneficial  for problems appearing in the field of electrical distribution systems  especially for the optimization of self-adequacy of interconnected microgrids and other related problems.


\end{document}